\pdfoutput=1

\documentclass[11pt]{article}

\usepackage[]{acl}

\usepackage{times}
\usepackage{latexsym}

\usepackage[T1]{fontenc}

\usepackage[utf8]{inputenc}

\usepackage{microtype}

\usepackage{graphicx}
\usepackage[utf8]{inputenc} 
\usepackage{booktabs}       
\usepackage{amsfonts}       
\usepackage{nicefrac}       
\usepackage{microtype}      
\usepackage{xcolor}         
\usepackage{multirow}
\usepackage{makecell}
\usepackage{mathtools}
\usepackage{lipsum}
\usepackage[flushleft]{threeparttable}
\usepackage{kotex}

\title{Hypergraph Transformer: Weakly-Supervised Multi-hop Reasoning \\ for Knowledge-based Visual Question Answering}

\author{Yu-Jung Heo\textsuperscript{\rm 1,4}, 
    Eun-Sol Kim\textsuperscript{\rm 2}, 
    Woo Suk Choi\textsuperscript{\rm 1},  
    and Byoung-Tak Zhang\textsuperscript{\rm 1,3} \\
    \textsuperscript{\rm 1} Seoul National University    
    \textsuperscript{\rm 2} Department of Computer Science, Hanyang University \\
    \textsuperscript{\rm 3} AI Institute (AIIS), Seoul National University
    \textsuperscript{\rm 4} Surromind\\
    yjheo@bi.snu.ac.kr, eunsolkim@hanyang.ac.kr, \{wschoi, btzhang\}@bi.snu.ac.kr
}

\begin{document}
\maketitle
\begin{abstract}
Knowledge-based visual question answering (QA) aims to answer a question which requires visually-grounded external knowledge beyond image content itself. 
Answering complex questions that require multi-hop reasoning under weak supervision is considered as a challenging problem since i) no supervision is given to the reasoning process and ii) high-order semantics of multi-hop knowledge facts need to be captured.
In this paper, we introduce a concept of hypergraph to encode high-level semantics of a question and a knowledge base, and to learn high-order associations between them. The proposed model, Hypergraph Transformer, constructs a question hypergraph and a query-aware knowledge hypergraph, and infers an answer by encoding inter-associations between two hypergraphs and intra-associations in both hypergraph itself. Extensive experiments on two knowledge-based visual QA and two knowledge-based textual QA demonstrate the effectiveness of our method, especially for multi-hop reasoning problem. Our source code is available at \url{https://github.com/yujungheo/kbvqa-public}.
\end{abstract}

\section{Introduction}
Visual question answering (VQA) is a semantic reasoning task that aims to answer questions about visual content depicted in images~\cite{antol2015vqa,zhu2016visual7w,hudson2019gqa}, and has become one of the most active areas of research with advances in natural language processing and computer vision. 
Recently, researches for VQA have advanced, from inferring visual properties on entities in a given image, to inferring commonsense or world knowledge about those entities~\cite{wang2017kbvqa,wang2018fvqa,marino2019okvqa,shah2019kvqa,zellers2019recognition}.

\begin{figure}[t]
\centering
    \includegraphics[width=1.0\columnwidth]{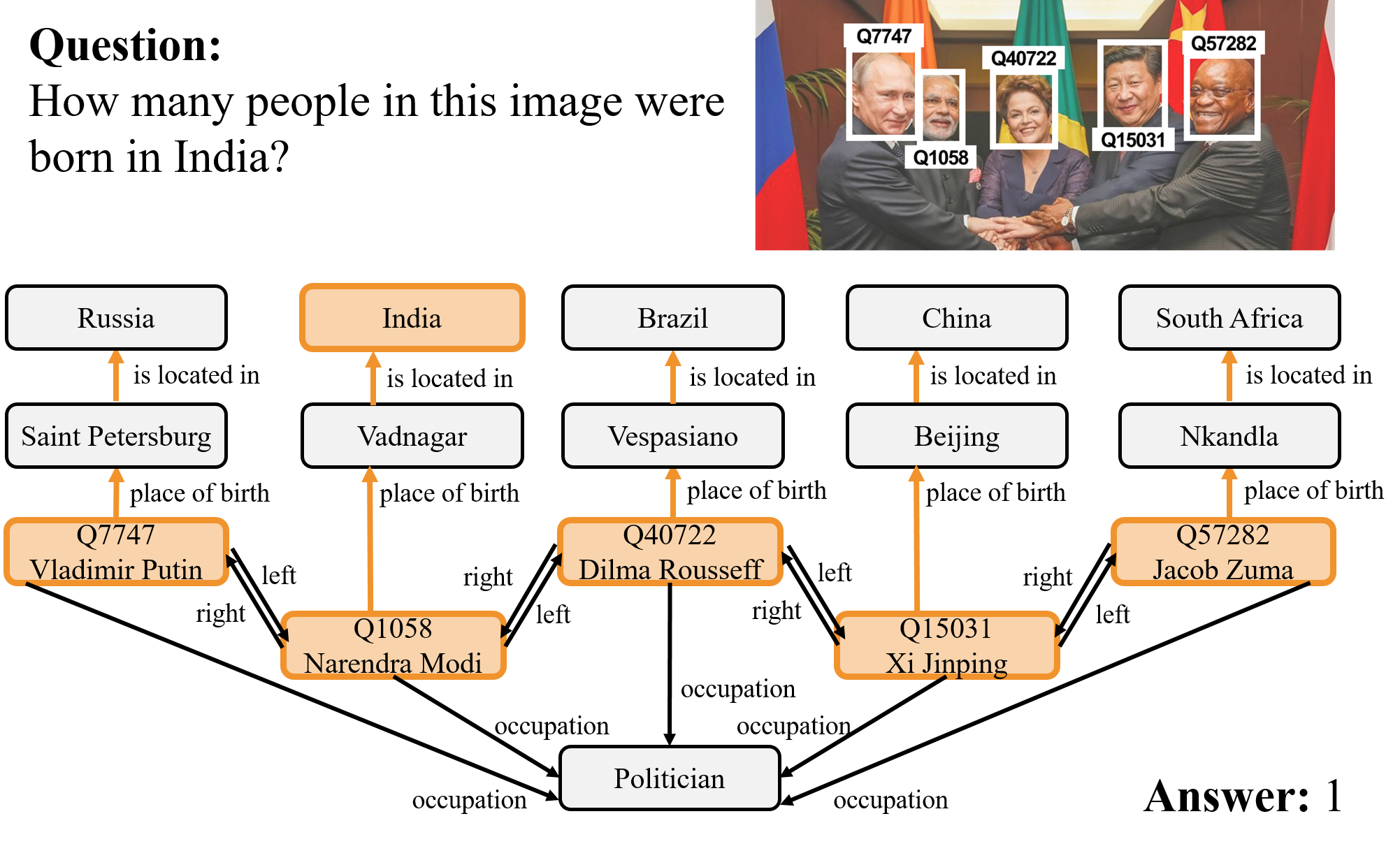}
\caption{An example of knowledge-based visual question answering. The rectangles and arrows between the rectangles represent the entities and relations from KB. To answer the given question, the multiple reasoning evidences (marked as orange) are required.}
\label{fig:task}
\end{figure}

In this paper, we focus on the task which is called knowledge-based visual question answering, where a massive number of knowledge facts from a general knowledge base (KB) is given with an image-question pair. To answer the given question as shown in Figure \ref{fig:task}, a model should understand the semantics of the given question, link visual entities appearing in the given image to the KB, extract a number of evidences from the KB and predict an answer by aggregating semantics of both the question and the extracted evidences. Following these, there are two fundamental challenges in this task. i) To answer a complex question, multi-hop reasoning over multiple knowledge evidences is necessary. ii) Learning a complex reasoning process is difficult especially in a condition where only QA is provided without extra supervision on how to capture any evidence from the KB and infer based on them. That is, the model should learn which knowledge facts to be attended to and how to combine them to infer the correct answer on its own.
Following the previous work~\cite{zhou2018pqpql}, we call this setting \textit{under weak supervision}.

Under weak supervision, previous studies proposed memory-based methods~\cite{narasimhan2018straight,shah2019kvqa} and graph-based methods~\cite{Narasimhan18OOB,zhu2020mucko} to learn to selectively focus on necessary pieces of knowledge.
The memory-based methods represent knowledge facts in a form of memory and calculate soft attention scores of each memory with respect to a question. Then, it infers an answer by attending to knowledge evidence with high attention scores.
On the other hand, to explicitly consider relational structure between knowledge facts, graph-based methods construct a query-aware knowledge graph by retrieving facts from KB and perform graph reasoning for a question. These methods mainly adopt an iterative message passing process to propagate information between adjacent nodes in the graph. However, it is difficult to capture multi-hop relationships containing long-distance nodes from the graph due to the well-known over-smoothing problem, where repetitive message passing process to propagate information across long distance makes features of connected nodes too similar and undiscriminating ~\cite{li2018oversmoothing,wang2020longdistance}. 

To address the above limitation, we propose a novel method, Hypergraph Transformer, which exploits hypergraph structure to encode multi-hop relationships and transformer-based attention mechanism to learn to pay attention to important knowledge evidences for a question.
We construct a question hypergraph and a knowledge hypergraph to explicitly encode high-order semantics present in the question and each knowledge fact, and capture multi-hop relational knowledge facts effectively. Then, we perform hyperedge matching between the two hypergraphs by leveraging transformer-based attention mechanism. We argue that introducing the concept of hypergraph is powerful for multi-hop reasoning problem in that it can encode high-order semantics without the constraint of length and learn cross-modal high-order associations.

The main contributions of this paper can be summarized as follows. i) We propose Hypergraph Transformer which enhances multi-hop reasoning ability by encoding high-order semantics in the form of a hypergraph and learning inter- and intra- high-order associations in hypergraphs using the attention mechanism.
ii) We conduct extensive experiments on two knowledge-based VQA datasets (KVQA and FVQA) and two knowledge-based textual QA datasets (PQ and PQL) and show superior performances on all datasets, especially multi-hop reasoning problem. iii) We qualitatively observe that Hypergraph Transformer performs robust inference by focusing on correct reasoning evidences under weak supervision. 

\begin{figure*}[t]
\centering
\includegraphics[width=1.0\textwidth]{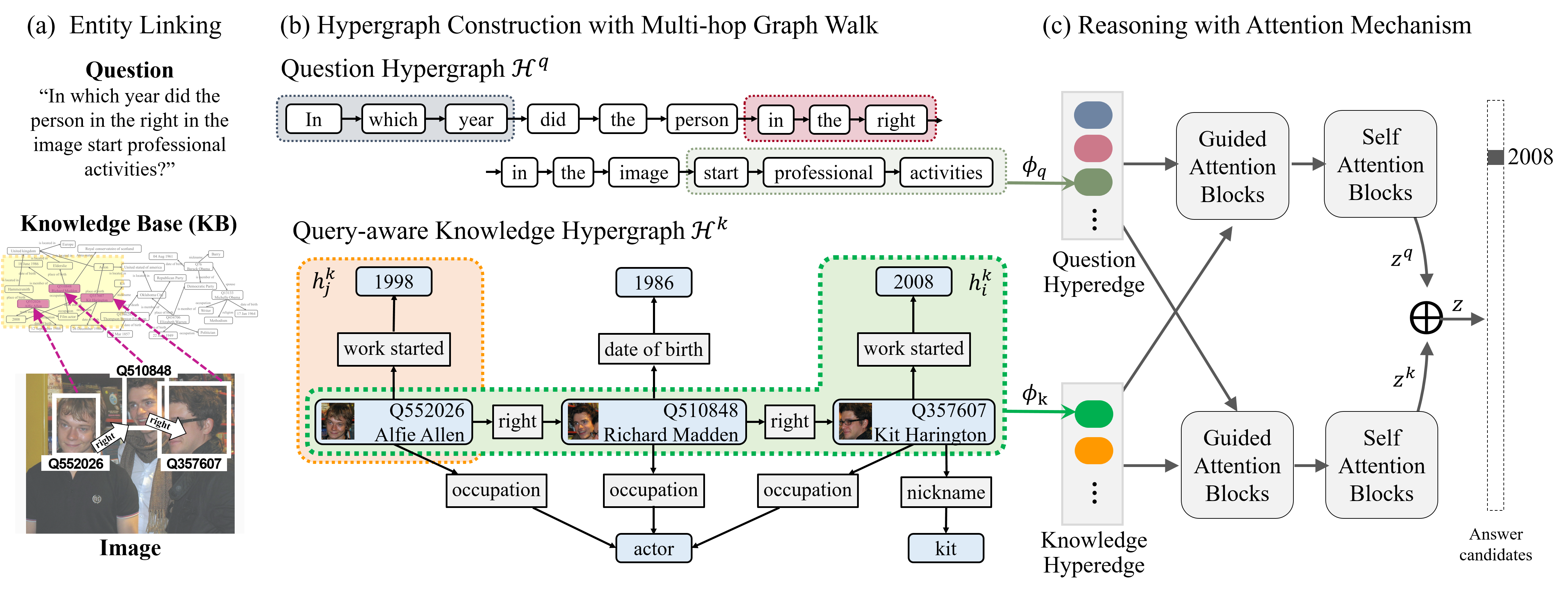}
\caption{The overview of Hypergraph Transformer. (a) Entity linking module links concepts from query (a given image and a question) to KB. (b) Query-aware knowledge hypergraph $\mathcal{H}^k$ and question hypergraph $\mathcal{H}^q$ are constructed by multi-hop graph walk. (c) Two hyperedge sets are fed into the guided-attention and self-attention blocks to learn inter- and intra-association in them. The joint representation is used to predict an answer.}
\label{fig:method}
\end{figure*}

\section{Related Work}
\paragraph{Knowledge-based visual question answering} 
\cite{wang2017kbvqa,wang2018fvqa,shah2019kvqa,marino2019okvqa,VLQAchallenge} proposed benchmark datasets for knowledge-based visual question answering that requires reasoning about an image on the basis of facts from a large-scale knowledge base (KB) such as Freebase~\cite{bollacker2008freebase} or DBPedia~\cite{auer2007dbpedia}.
To solve the task, two pioneering studies~\cite{wang2017kbvqa,wang2018fvqa} suggested logical parsing-based methods which convert a question to a KB logic query using pre-defined query templates and execute the generated query on KB for searching an answer. Since then information retrieval-based methods which retrieve knowledge facts associated with a question and conduct semantic matching between the facts and the question are introduced. \cite{narasimhan2018straight,shah2019kvqa} proposed memory-based methods that represent knowledge facts in the form of memory and calculate soft attention scores of the memory with a question. \cite{Narasimhan18OOB,zhu2020mucko} represented the retrieved facts as a graph and performed graph reasoning through message passing scheme utilizing graph convolution. However, these methods are complicated to encode inherent high-order semantics and multi-hop relationships present in the knowledge graph. Therefore, we introduce a concept of hypergraph and propose transformer-based attention mechanism over hypergraphs.

\paragraph{Multi-hop knowledge graph reasoning} is a process of sequential reasoning based on multiple evidences of a knowledge graph, and has been broadly used in various downstream tasks such as question answering~\cite{lin2019kagnet,saxena2020embedkgqa,han2020hgcqa,han2020open,yadati2021recursivehg}, or knowledge-enhanced text generation~\cite{liu2019knowledge, moon2019opendialkg,ji2020langgen}. 
Recent researches have introduced the concept of hypergraph for multi-hop graph reasoning~\cite{kim2020hypergraph,han2020hgcqa,han2020open,yadati2019hypergcn,yadati2021recursivehg,sun2020knowledge}. These models have a similar motivation to the Hypergraph Transformer proposed in this paper, but core operations are vastly different.
These models mainly update node representations in the hypergraph through a message passing process using graph convolution operation. On the contrary, our method update node representations via hyperedge matching of hypergraphs instead of message passing scheme. We argue that this update process effectively learns the high-order semantics inherent in each hypergraph and the high-order associations between two hypergraphs.

\section{Method}

\subsection{Notation}
To capture high-order semantics inherent in the knowledge sources, we adopt the concept of hypergraph. Formally, directed hypergraph $\mathcal{H} = \{\mathcal{V}, \mathcal{E}\}$ is defined by a set of nodes $\mathcal{V}=\{v_1, ..., v_{|\mathcal{V}|}\}$ and a set of hyperedges $\mathcal{E}=\{h_1, ... , h_{|\mathcal{E}|}\}$. Each node is represented as a $w$-dimensional embedding vector, i.e., $v_i \in \mathbb{R}^w$. Each hyperedge connects an arbitrary number of nodes and has partial order itself,
i.e., $h_i = \{v'_1 \preceq ... \preceq v'_l\}$ where $\mathcal{V'}=\{v'_1, ..., v'_l\} \text{ is a subset of } \mathcal{V}$ and $\preceq$ is a binary relation which denotes an element ($v'_i$) precedes the other ($v'_j$) in the ordering when $v'_i \preceq v'_j$. A hyperedge is flexible to encode different kinds of semantics in the underlying graph without the constraint of length.

\subsection{Entity linking}\label{entlink}
As shown in Figure \ref{fig:method}(a), entity linking module first links concepts from query (a given image-question pair) to knowledge base. We detect visual concepts (e.g., objects, attributes, person names) in a given image and named entities in a given question. The semantic labels of visual concepts or named entities are then linked with knowledge entities in the knowledge base using exact keyword matching.

\subsection{Hypergraph construction}
\paragraph{Query-aware knowledge hypergraph} 
A knowledge base (KB), a vast amount of general knowledge facts, contains not only knowledge facts required to answer a given question but also unnecessary knowledge facts. Thus, we construct a query-aware knowledge hypergraph $\mathcal{H}^k=\{\mathcal{V}^k, \mathcal{E}^k\}$ to extract related information for answering a given question. It consists of a node set $\mathcal{V}^k$ and hyperedge set $\mathcal{E}^k$, which represent a set of entities in knowledge facts and a set of hyperedges, respectively. Each hyperedge connects the subset of vertices $\mathcal{V}'^k \subset \mathcal{V}^k$.

We consider a huge number of knowledge facts in the KB as a huge knowledge graph, and construct a hypergraph by traversing the knowledge graph.
Such traversal, called graph walk, starts from the node linked from the previous module (see section \ref{entlink}) and considers all entity nodes associated with the start node. 
We define a triplet as a basic unit of graph walk to preserve high-order semantics inherent in knowledge graph, i.e., every single graph walk contains three nodes \textit{\{head, predicate, tail\}}, rather than having only one of these three nodes.
In addition to the triplet-based graph walks, a multi-hop graph walk is proposed to encode multiple relational facts that are interconnected.
Multi-hop graph walk connects multiple facts by setting the arrival node (\textit{tail}) of the preceding walk as the starting (\textit{head}) node of the next walk, thus, $n$-hop graph walk combines $n$ facts as a hyperedge.

\paragraph{Question hypergraph} We transform a question sentence into a question hypergraph $\mathcal{H}^q$ consisting of a node set $\mathcal{V}^q$ and a hyperedge set $\mathcal{E}^q$. We assume that each word unit (a word or named entity) of the question is defined as a node, and has edges to adjacent nodes. For question hypergraph, each word unit is used as a start node of a graph walk. The multi-hop graph walk is conducted in the same manner as the knowledge hypergraph. A $n$-gram phrase is considered as a hyperedge in the question hypergraph (see Figure \ref{fig:method}(b)).

\subsection{Reasoning with attention mechanism}
To consider high-order associations between knowledge and question, we devise structural semantic matching between the query-aware knowledge hypergraph and the question hypergraph. We introduce an attention mechanism over two hypergraphs based on guided-attention~\cite{tsai2019multimodal} and self-attention~\cite{vaswani2017attention}. As shown in Figure \ref{fig:method}(c), the guided-attention blocks are introduced to learn correlations between knowledge hyperedges and question hyperedges by inter-attention mechanism, and then intra-relationships of in knowledge or question hyperedges are trained with the following self-attention blocks. The details of two modules, guided-attention blocks and self-attention blocks, are described as below. Note that we use $Q$, $K$, and $V$ for query, key, value, and $q$, $k$ as subscripts to represent question and knowledge, respectively.

\paragraph{Guided-attention} 
To learn inter-association between two hypergraphs, we first embed a knowledge hyperedge and a question hyperedge as follows: $e^k = \phi_k \circ f_k(h^k) \in \mathbb{R}^d, e^q = \phi_q  \circ f_q(h^q) \in \mathbb{R}^d$ where $h^{[\cdot]}$ is a hyperedge in $\mathcal{E}^{[\cdot]}$. Here, $f_{[\cdot]}$ is a hyperedge embedding function and $\phi_{[\cdot]}$ is a linear projection function. The design and implementation of $f_{[\cdot]}$ are not constrained (e.g., any pooling operation or any learnable neural networks), but we use a simple concatenation operation of node representations in a hyperedge as $f_{[\cdot]}$. The representations of hyperedges in the same hypergraph (e.g., $e^k, e^q$) are packed together into a matrix $E^k$ and $E^q$.

We define the knowledge hyperedges $E^k$ and the question hyperedges $E^q$ as a query and key-value pairs, respectively. We set a query $Q_{k}=E^kW_{Q_{k}}$, a key $K_{q}=E^qW_{K_{q}}$, and a value $V_{q}=E^qW_{V_{q}}$, where all projection matrices $W_{[\cdot]} \in \mathbb{R}^{d \times d_v}$ are learnable parameters.
Then, scaled dot product attention using the query, key, and value is calculated as $\operatorname{Attention}(Q_k, K_q, V_q) = \mathrm{softmax}(\frac{Q_k K_q^T}{\sqrt{d_v}})V_q$ where $d_v$ is the dimension of the query and the key vector. In addition, the guided-attention which uses the question hyperedges as query and the knowledge hyperedges as key-value pairs is performed in a similar manner: $\operatorname{Attention}(Q_q, K_k, V_k)$.

\paragraph{Self-attention}
The only difference between guided-attention and self-attention is that the same input is used for both query and key-value within self-attention. For example, we set query, key, and value based on the knowledge hyperedges $E_k$, and the self-attention for knowledge hyperedges is conducted by $\operatorname{Attention}(Q_k, K_k, V_k)$. For question hyperedges $E_q$, self-attention is performed in a similar manner: $\operatorname{Attention}(Q_q, K_q, V_q)$.

Following the standard structure of the transformer, we build up guided-attention block and self-attention block where each block consists of each attention operation with layer normalization, residual connection, and a single feed-forward layer. By passing the guided-attention blocks and self-attention blocks sequentially, representations of knowledge hyperedges and question hyperedges are updated and finally aggregated to single vector representation as $z_k \in \mathbb{R}^{d_v}$ and $z_q \in \mathbb{R}^{d_v}$, respectively.

\begin{table*}[t]
\centering
\begin{tabular}{lccccccc}
\Xhline{2\arrayrulewidth}
\multirow{2}{*}{Model} & \multicolumn{3}{c}{Original (ORG)} & \multicolumn{3}{c}{Paraphrased (PRP)} & \multirow{2}{*}{Mean} \\
 & 1-hop & 2-hop & 3-hop & 1-hop & 2-hop & 3-hop &\\
\hline
BLSTM & - & - & - & - & - & - & 51.0 \\
MemNN~\cite{sukhbaatar2015end} & - & - & - & - & - & - & 59.2 \\
\hline
GCN~\cite{kipf2017semi} & 65.7 & 67.4 & 66.9 & 65.8 & 67.5 & 67.0 & 66.7 \\
GGNN~\cite{ggnn} & 72.9 & 74.5 & 74.0 & 72.9 & 74.6 & 74.1 & 73.8 \\ 
MemNN$\dagger$~\cite{sukhbaatar2015end} & 78.1 & 77.8 & 76.1 & 78.0 & 78.1 & 76.0 & 77.3 \\ 
HAN~\cite{kim2020hypergraph} & 77.5 & 77.5 & 77.2 & 77.1 & 77.4 & 76.9 & 77.3 \\
BAN~\cite{kim2018bilinear} & 83.5 & 84.0 & 83.7 & 83.7 & 84.3 & 83.8 & 83.8 \\
\hline
\textbf{Ours} & \textbf{88.1} & \textbf{90.2} & \textbf{91.0} & \textbf{87.8} & \textbf{90.5} & \textbf{90.7} & \textbf{89.7} \\
\Xhline{2\arrayrulewidth}
\end{tabular}
\caption{QA accuracy on oracle setting in KVQA under weak supervision. ORG and PRP are a type of question and 1-hop, 2-hop, and 3-hop are the number of graph walks to construct a knowledge hypergraph. The performance of BLSTM and MemNN is reported in~\cite{shah2019kvqa} and we re-implemented MemNN$\dagger$ for a fair comparison.}
\label{table:orc}
\end{table*}

\subsection{Answer predictor}
To predict an answer, we first concatenate the representation $z_k$ and $z_q$ obtained from the attention blocks and feed into a single feed-forward layer (i.e., $\mathbb{R}^{2d_v} \mapsto \mathbb{R}^w$) to make a joint representation $z$. We then consider two types of answer predictor: multi-layer perceptron and similarity-based answer predictor. Multi-layer perceptron as an answer classifier $p=\psi(z)$ is a prevalent for visual question answering problems. For similarity-based answer, we calculate a dot product similarity $p=z C^T$ between $z$ and answer candidate set $C \in \mathbb{R}^{|\mathcal{A}| \times w}$ where $|\mathcal{A}|$ is a number of candidate answers and $w$ is a dimension of representation for each answer. The most similar answer to the joint representation is selected 
as an answer among the answer candidates. For training, we use only supervision from QA pairs without annotations for ground-truth reasoning paths.
To this end, cross-entropy between prediction $p$ and ground-truth $t$ is utilized as a loss function.

\section{Experimental Settings}
\subsection{Datasets}
In this paper, we evaluate our model across various benchmark datasets: Knowledge-aware VQA (KVQA) ~\cite{shah2019kvqa}, Fact-based VQA (FVQA)~\cite{wang2018fvqa}, PathQuestion (PQ) and PathQuestion-Large (PQL)~\cite{zhou2018pqpql}. 
KVQA, a large-scale benchmark dataset for complex VQA, contains 183,007 pairs for 24,602 images from Wikipedia and corresponding captions, and provides 174,006 knowledge facts for 39,414 unique named entities based on Wikidata~\cite{vrandevcic2014wikidata} since it requires world knowledge beyond visual content.
KVQA consists of two types of questions: original (ORG) and paraphrased (PRP) question generated from the original question via the online paraphrasing tool. 
FVQA, a representative dataset for commonsense-enabled VQA, considers external knowledge about common nouns depicted in a given image, and contains 5,826 QA pairs for 2,190 images and 4,216 unique knowledge facts from DBPedia~\cite{auer2007dbpedia}, ConceptNet~\cite{liu2004conceptnet}, and WebChild~\cite{tandon2014webchild}.
The last two datasets, PQ and PQL, focus on evaluating multi-hop reasoning ability in the knowledge-based textual QA task. PQ and PQL contain 7,106 and 2,625 QA pairs on 4,050 and 9,844 knowledge facts from the subset of Freebase~\cite{bollacker2008freebase}, respectively. The detailed statistics of the datasets are shown in Appendix \ref{ax:datastat}. 

\subsection{Implementation details}
Each node in the knowledge hypergraph and the question hypergraph is represented as a 300-dimensional vector (i.e., $w=300$) initialized using GloVe~\cite{pennington2014glove}. 
Random initialization is applied when a word for a node does not exist in the vocabulary of GloVe. Mean pooling is applied when a node consists of multiple words.
For entity linking for KVQA, we apply the well-known pre-trained models for face identification: RetinaFace~\cite{deng2020retinaface} for face detection and ArcFace~\cite{Deng_2019_CVPR} for face feature extraction. 
For all datasets, we follow the experimental settings as in previous works. 
We use the similarity-based answer predictor for KVQA, and MLP for the others.
We adopt Adam~\cite{iclr15adam} to optimize all learnable parameters in the model. We describe details of the experimental settings and the tuned hyperparameters for each dataset in Appendix \ref{ax:expdetails}.

\begin{table*}[]
\centering
\begin{threeparttable}
\begin{tabular}{l|ccc|ccc}
\Xhline{2\arrayrulewidth}
 & \multicolumn{3}{c|}{PathQuestion} & \multicolumn{3}{c}{PathQuestion-Large} \\
  & PQ-2H & PQ-3H & PQ-M & PQL-2H & PQL-3H & PQL-M \\
\hline
Seq2Seq~\cite{NIPS2014_a14ac55a} & 89.9 & 77.0 & - & 71.9 & 64.7 & - \\
MemNN~\cite{sukhbaatar2015end} & 89.5 & 79.2 & 86.8 & 61.2 & 53.6 & 55.8 \\
KV-MemNN~\cite{miller2016key} & 91.5 & 79.4 & 85.2 & 70.5 & 63.4 & 68.6 \\
IRN~\cite{zhou2018pqpql} & 96.0 & 87.7 & - & 72.5 & 71.0 & - \\
\hline
Embed~\cite{bordes2014open} & 78.7 & 48.3 & - & 42.5 & 22.5 & - \\
Subgraph~\cite{bordes2014question} & 74.4 & 50.6 & - & 50.0 & 21.3 & - \\
MINERVA~\cite{das2018go} & 75.9 & 71.2 & 73.1 & 71.8 & 65.7 & 66.9 \\
IRN-weak~\cite{zhou2018pqpql} & 91.9 & 83.3 & 85.8 & 63.0 & 61.8 & 62.4 \\
SRN~\cite{qiu2020stepwise} & 96.3 & 89.2 & 89.3 & 78.6 & 77.5 & 78.3 \\ 
\textbf{Ours} & \textbf{96.4} & \textbf{90.3} & \textbf{89.5} & \textbf{90.5} & \textbf{77.9}(*) & \textbf{94.5} \\
\Xhline{2\arrayrulewidth}
\end{tabular}
\begin{tablenotes}
    \small
    \item (*) For PQL-3H-More data (2x QA pairs on the same KB as PQL-3H), our model shows 95.4\% accuracy. 
\end{tablenotes}
\end{threeparttable}
\caption{Accuracy on PathQuestion (PQ) and PathQuestion-Large (PQL). 2H and 3H represent the number of multi-hops in ground-truth reasoning paths to answer given questions, and M represents the mixture of 2H and 3H. The models in the first block employ a ground-truth reasoning path as extra supervision (i.e., fully-supervised), and the models in the second block including our model are under weak supervision.}
\label{table:pqpql}
\end{table*}

\section{Quantitative Results}
\subsection{Knowledge-aware visual question answering}

We compare the proposed model, Hypergraph Transformer, with other comparative state-of-the-art methods. We report performances on original (ORG) and paraphrased (PRP) questions according to the number of graph walk. 
For comparative models, three kinds of methods are considered, which are graph-based, memory-based and attention-based networks. The detailed description about the comparative models is described in Appendix \ref{ax:impdetailsKVQA}. 
To evaluate a pure reasoning ability of the models regardless of the performance of entity linking, we first conduct experiments in the oracle setting which ground-truth named entities in an image are given.

As shown in Table \ref{table:orc}, our model outperforms comparative models with a large margin across all settings.
From the results, we find that the attention mechanism between question and knowledge is crucial for complex QA. Since GCN~\cite{kipf2017semi} and GGNN~\cite{ggnn} encode question and knowledge graph separately, they do not learn interactions between question and knowledge. Thus, GCN and GGNN show quite low performance under 74\% mean accuracy. On the other hand, MemNN$\dagger$~\cite{memnet}, HAN~\cite{kim2020hypergraph}, and BAN~\cite{kim2018bilinear} achieve comparatively high performance because MemNN$\dagger$ adopts question-guided soft attention over knowledge memories. HAN and BAN utilize multi-head co-attention between question and knowledge. 

\paragraph{Entity linking setting}
We also present the experimental results on the entity linking setting where the named entities are not provided as the oracle setting, but detected by the module as described in Section \ref{entlink}.
As shown in Table \ref{table:det} of Appendix \ref{ax:impdetailsKVQA}, our model shows the best performances for both original and paraphrased questions. For all comparative models, we use the same knowledge hypergraph extracted by the 3-hop graph walk. In entity linking setting, the constructed knowledge hypergraph can be incomplete and quite noisy due to the undetected entities or misclassified entity labels. However, Hypergraph Transformer shows robust reasoning capacity over the noisy inputs. Here, we remark that the upper bound of QA performance is 72.8\% due to the error rate of entity linking module. We expect that the performance will be improved when the entity linking module is enhanced.

\begin{table*}[t]
\centering
\begin{tabular}{l|cc|ccc|ccc|cc}
\Xhline{2\arrayrulewidth}
 \multirow{2}{*}{Model} & \multicolumn{2}{c|}{Inputs} & \multicolumn{3}{c|}{Original (ORG)} & \multicolumn{3}{c|}{Paraphrased (PRP)} & \multirow{2}{*}{Mean} \\
 & Knowledge & Question & 1-hop & 2-hop & 3-hop & 1-hop & 2-hop & 3-hop &\\
\hline
 (a) SA & Word & Word & 79.4 & 79.6 & 77.6 & 77.1 & 77.7 & 77.7 & 78.2 \\
 (b) SA+GA & Word & Word & 80.9 & 82.3 & 81.5 & 80.7 & 82.2 & 81.8 &81.6 \\
 (c) SA+GA & Word & Hyperedge & 82.1 & 84.2 & 82.8 & 81.1 & 83.5 & 82.3 & 82.7\\
 (d) SA+GA & Hyperedge & Word & 87.0 & 89.9 & 88.9 & 87.3 & 89.7 & 89.2 & 88.7\\
 \hline
 (e) SA+GA & \multirow{2}{*}{Hyperedge} & \multirow{2}{*}{Hyperedge} & \multirow{2}{*}{\textbf{88.1}} & \multirow{2}{*}{\textbf{90.2}} & \multirow{2}{*}{\textbf{91.0}} & \multirow{2}{*}{\textbf{87.8}} & \multirow{2}{*}{\textbf{90.5}} & \multirow{2}{*}{\textbf{90.7}} & \multirow{2}{*}{\textbf{89.7}} \\
 \textbf{$ \ \ \ \ \ $(Ours)} & & & & & & & & & \\
  \hline
 (f) \textbf{Ours}-SA & Hyperedge & Hyperedge & 85.2 & 88.8 & 88.3 & 85.0 & 88.3 & 88.4 & 87.1 \\ 
 (g) \textbf{Ours}-GA & Hyperedge & Hyperedge & 82.6 & 83.6 & 85.0 & 82.7 & 83.6 & 84.9 & 83.7 \\
\Xhline{2\arrayrulewidth}
\end{tabular}
\caption{(a-e) Validation for the effectiveness of using hypergraph. Here, we compare the results with respect to the different types of the input format (i.e., Single Word or Hyperedge) used to represent knowledge and question which are fed into the attention mechanism. (e-g) Ablation study for attention blocks of Hypergraph Transformer. GA and SA are abbreviations of guided-attention and self-attention, respectively.}
\label{table:HGTvalidation}
\end{table*}

\subsection{Fact-based visual question answering}
We conduct experiments on Fact-based Visual Question Answering (FVQA) as an additional benchmark dataset for knowledge-based VQA.
Different from KVQA focusing on world knowledge for named entities, FVQA considers commonsense knowledge about common nouns in a given image. Here, we assume that the performance of entity linking is perfect, and evaluate the pure reasoning ability of our model. As shown in Table \ref{table:fvqa} of Appendix \ref{ax:expdetails}, Hypergraph Transformer shows comparable performance in both top-1 and top-3 accuracy in comparison with the state-of-the-art methods. We confirm that our model works effectively as a general reasoning framework without considering characteristics of different knowledge sources (i.e., Wikidata for KVQA, DBpedia, ConceptNet, WebChild for FVQA).

\subsection{PathQuestion and PathQuestion-Large}
To verify multi-hop reasoning ability of our model, we conduct experiments on PathQuestion (PQ) and PathQuestion-Large (PQL). PQ and PQL datasets have annotations of a ground-truth reasoning path to answer a given question. Specifically, \{PQ, PQL\}-\{2H, 3H\} denotes a split of PQ and PQL with respect to the number of hops in ground-truth reasoning paths (i.e., 2-hop or 3-hop). \{PQ, PQL\}-M is a mixture of the 2-hop and 3-hop questions in both dataset, and used to evaluate the more general scenario where the number of reasoning path required to answer a given question is unknown. 

The experimental results on diverse split of PQ and PQL datasets are provided in Table \ref{table:pqpql}. The first section in the table includes fully-supervised models which require a ground-truth path annotation as an additional supervision. The second section contains weakly-supervised models learning to infer the multi-hop reasoning paths without the ground-truth path annotation. Hypergraph Transformer is involved in the weakly-supervised models because it only exploits an answer as a supervision. 

Our model shows comparable performances on PQ-\{2H, 3H, M\} to the state-of-the-art weakly-supervised model, SRN. Especially, Hypergraph Transformer shows significant performance improvement (78.6\% $\rightarrow$ 90.5\% for PQL-2H, 78.3\% $\rightarrow$ 94.5\% for PQL-M) on PQL. We highlight that PQL is more challenging dataset than PQ in that PQL not only covers more knowledge facts but also has fewer QA instances.
We observe that the accuracy on PQL-3H is relatively lower than the other splits. This is due to the insufficient number of training QA pairs in PQL-3H. When we use PQL-3H-More which has twice more QA pairs (1031 $\rightarrow$ 2062) on the same knowledge base as PQL-3H, our model achieves 95.4\% accuracy.

\section{Validation for Hypergraph Transformer}
We verify the effectiveness of each module in Hypergraph Transformer. To analyze the performances of the variants in our model, we use KVQA which is a representative and large-scale dataset for knowledge-based VQA. Here, we mainly focus on two aspects: i) effect of hypergraph and ii) effect of attention mechanism. To evaluate a pure reasoning ability of the models, we conduct experiments in the oracle setting.

\begin{figure*}
  \centering
  \includegraphics[width=1.0\textwidth]{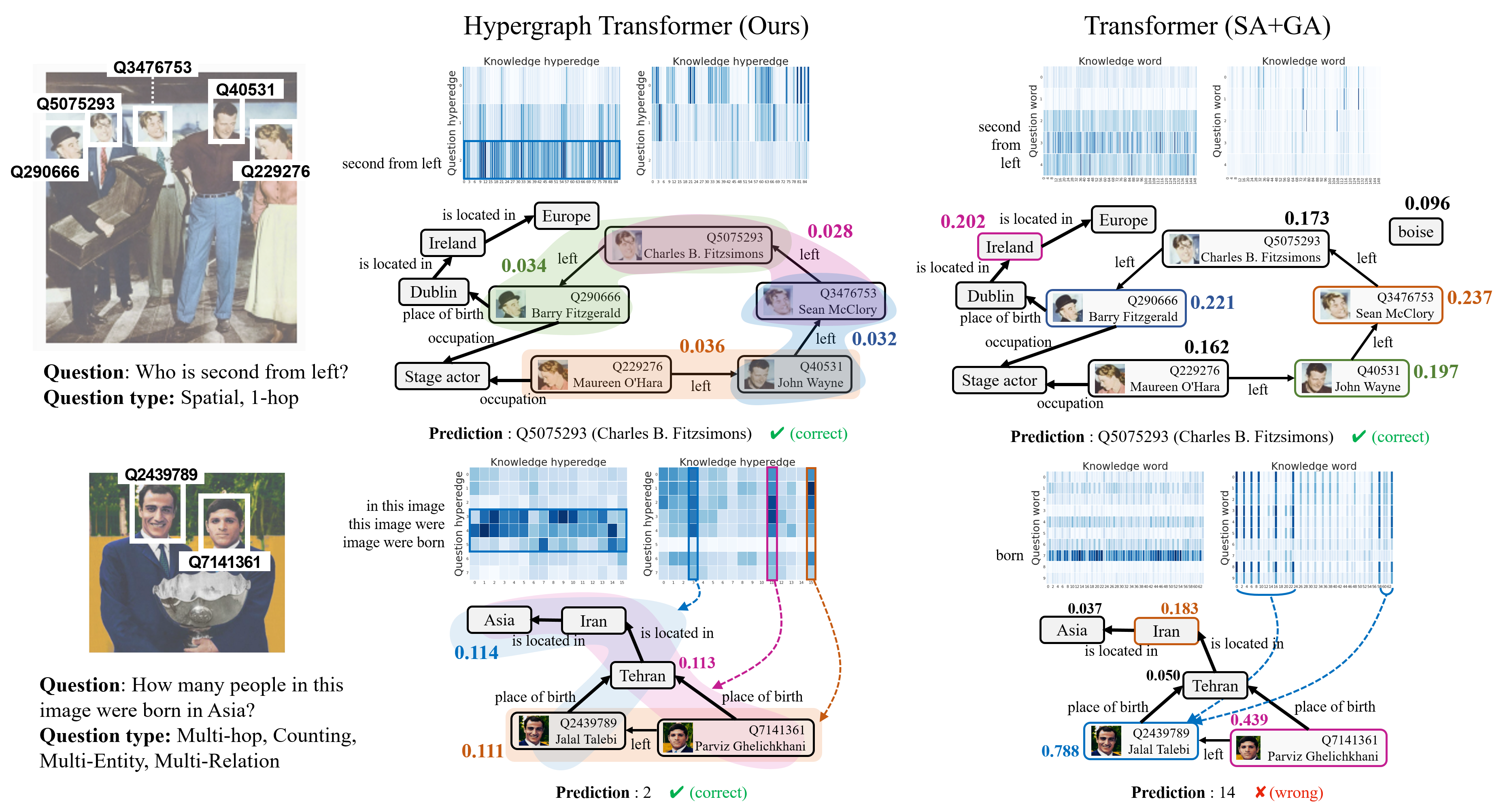}
  \caption{Qualitative analysis on effectiveness of using hypergraph as input format to Transformer architecture. Here, we visualize attention maps for Hypergraph Transformer and the Transformer (SA+GA). All attention scores are averaged over multi-heads and multi-layers. Each $x$ and $y$ axis represent indices of question and knowledge hyperedges in Hypergraph Transformer, and indices of question and knowledge word in Transformer (SA+GA). In the attention maps, the dark colors represent high values. The hyperedges with high attention scores are visualized.}
  \label{fig:kvqa_atts}
\end{figure*}

\subsection{Effect of hypergraph}
To analyze the effectiveness of hypergraph-based input representation, we conduct comparative experiments on the different types of input formats for Transformer architecture. Here, we consider the two types of input format, which are single-word-unit and hyperedge-based representations. Compared to hyperedge-based inputs considering multiple relational facts as a input token, single-word-unit takes every entity and relation tokens as separate input tokens.
We note that using single-word-unit-based input format for both knowledge and question is the standard settings for the Transformer network and using hyperedge-based input format for both is the proposed model, Hypergraph Transformer. We set the Transformer (SA+GA) as a backbone model, and present the results in Table \ref{table:HGTvalidation}(b-e).
When hypergraph-based representations are used for both knowledge and question, the results show the best performance across all settings over question types (ORG and PRP) and a number of graph walk (1-hop, 2-hop, and 3-hop). As shown in Table \ref{table:HGTvalidation}, the mean accuracy of QA achieves 89.7\% when both are encoded using hyperedges, while using single-word-unit-based representation causes performance to drop to 81.6\%. Especially, when we convert the one of both hyperedge-level representation to single-word-unit-based representation, the mean accuracy of QA is 82.7\% and 88.7\%, respectively. These results validate that it is meaningful to consider not only knowledge but also question as hypergraphs. 

\paragraph{Effect of multi-hop graph walk} We compare the performances with different number of graph walks used to construct a knowledge hypergraph (i.e., 1-hop, 2-hop, and 3-hop).
All models except ours show slightly lower performance on the 3-hop graph than on the 2-hop graph. We observe that the number of extracted knowledge facts increases when the number of graph walk increases, and unnecessary facts for answering a given question are usually included. Nonetheless, our model shows robust reasoning performance when a large and noisy knowledge facts are given. 

\subsection{Effect of attention mechanism} 
To investigate the impacts of each attention block (i.e., GA and SA), ablation studies are shown in Table \ref{table:HGTvalidation}(e-g).
The scores across all settings drop when GA or SA is removed. Particularly, the mean accuracy of QA is decreased by 6.0\% (89.7\% $\rightarrow$ 83.7\%), 2.6\% (89.7\% $\rightarrow$ 87.1\%) for cutting out the GA and the SA block, respectively. Based on the two experiments, we identify that not only the guided-attention which captures inter-relationships between question and knowledge but also the self-attention which learns intra-relationship in them are crucial to the complex QA. 
To sum up, Hypergraph Transformer takes graph-level inputs, i.e., hyperedge, and conducts semantic matching between hyperedges by the attention mechanism. Due to the two characteristics, the model shows better reasoning performance focusing on the evidences necessary for reasoning under weak supervision. 

\section{Qualitative Analysis} 
Figure \ref{fig:kvqa_atts} provides the qualitative analysis on effectiveness of using a hypergraph as an input format to Transformer architecture. We present the attention map from the guided-attention block, and visualize top-$k$ attended knowledge facts or entities with the attention scores. In the first example, both model, Hypergraph Transformer and Transformer (SA+GA), infer the correct answer, \textit{Q5075293}. Our model responds by focusing on $\{\textit{second} \preceq \textit{from} \preceq \textit{left}\}$ phrase of the question and four facts having a \textit{left} relation among 86 knowledge hyperedges. In comparison, Transformer (SA+GA) strongly attends to the knowledge entities which appear repetitive in the knowledge facts. Especially, the model attends to \textit{Q3476753}, \textit{Q290666} and \textit{Ireland} with the high attention score 0.237, 0.221, and 0.202. In the second example, our model attends to the correct knowledge hyperedges considering the multi-hop facts about \textit{place of birth} of the people shown in the given image, and infers the correct answer. On the other hand, Transformer (SA+GA) strongly attends to the knowledge entity of person (\textit{Q2439789}) presented in the image with undesired attention score 0.788. The second and third attended knowledge entities are the other person (\textit{Q7141361}) and \textit{Iran}. Transformer (SA+GA) fails to focus on the multi-hop facts required to answer the given question and predicts the answer with the wrong number at the end. 

\section{Discussion and Conclusion}
In this paper, we proposed Hypergraph Transformer for multi-hop reasoning over knowledge graph under weak supervision. Hypergraph Transformer adopts hypergraph-based representation to encode high-order semantics of knowledge and questions and considers associations between a knowledge hypergraph and a question hypergraph. Here, each node representation in the hypergraphs is updated by inter- and intra-attention mechanisms in two hypergraphs, rather than by iterative message passing scheme. Thus, Hypergraph Transformer can mitigate the well-known over-smoothing problem in the previous graph-based methods exploiting the message passing scheme.
Extensive experiments on various datasets, KVQA, FVQA, PQ, and PQL validated that Hypergraph Transformer conducts accurate inference by focusing on knowledge evidences necessary for question from a large knowledge graph. Although not covered in this paper, an interesting future work is to construct heterogeneous knowledge graph that includes more diverse knowledge sources (e.g. documents on web).

\section*{Acknowledgements}
We would like to thank Woo Young Kang, Kyoung-Woon On, Seonil Son, Gi-Cheon Kang, Christina Baek, Junseok Park, Min Whoo Lee, Hwiyeol Jo and Sang-Woo Lee for their helpful comments and discussion. This work was partly supported by the IITP (2015-0-00310-SW.StarLab/20\%, 2017-0-01772-VTT/20\%, 2019-0-01371-BabyMind/10\%, 2021-0-02068-AIHub/10\%, 2021-0-01343-GSAI/10\%, 2020-0-01373/10\%) grants, the KIAT (P0006720-ILIAS/10\%) grant funded by the Korean government, and the Hanyang University (HY-202100000003160/10\%).

\bibliography{anthology,output}
\bibliographystyle{acl_natbib}

\newpage $ $
\newpage
\appendix

\paragraph{Appendix. } This supplementary material provides additional information not described in the main text due to the page limit. The contents of this appendix are as follows: In Section \ref{ax:datastat}, we show the detailed statistics for the diverse splits of four benchmark datasets, i.e., KVQA, FVQA, PQ and PQL. In Section \ref{ax:kvqa} and \ref{ax:pq}, we present the additional quantitative and qualitative analyses on KVQA and PQ datasets, respectively. In Section \ref{ax:expdetails}, we describe the experimental details for each dataset. In Section \ref{ax:impdetailsKVQA}, we depict the implementation details of comparative models for KVQA.

\section{Data Statistics}\label{ax:datastat}
The diverse split statistics for four benchmark datasets, KVQA~\cite{shah2019kvqa}, FVQA~\cite{wang2018fvqa}, PQ and PQL~\cite{zhou2018pqpql}, are shown in Table \ref{table:datastat}.
Here, we highlight four aspects as follows: 1) KVQA dataset covers the large number of entities (at least 5 times more) and knowledge facts (at least 17 times more) than FVQA, PQ and PQL. 2) PQ and PQL datasets have annotations of a ground-truth reasoning path to answer a given question. 2H and 3H denote the number of hops (i.e., 2-hop and 3-hop) in ground-truth reasoning paths. Also, M denotes a mixture of the 2H and 3H questions. 3) PQL covers more knowledge facts including a large number of entities and relations than PQ, but has fewer QA pairs. 4) PQL-3H has a quite limited number of QA pairs (1,031). PQL-3H-More has twice more QA pairs (2,062) with the same number of entities, relations, knowledge facts and answers as PQL-3H.

\begin{table*}[h]
\centering
\begin{threeparttable}
\begin{tabular}{lc|c|ccc|ccc}
\Xhline{2\arrayrulewidth}
 & KVQA & FVQA & PQ-2H & PQ-3H & PQ-M & PQL-2H & PQL-3H & PQL-M \\
\hline
\# Entities & 39,414 & 3,391 & 1,057 & 1,837 & 2,257 & 5,035 & 6,506 & 6,506 \\
\# Relations & 18 & 13 & 14 & 14 & 14 & 364 & 412 & 412 \\
\# Knowledge facts & 174,006 & 4,216 & 1,211 & 2,839 & 4,050 & 4,247 & 5,597 & 9,844 \\
\# Words & 63,164 & 6,663 & 1,180 & 1,929 & 2,407 & 5,505 & 7,001 & 7,034 \\
\# QA pairs & 183,007 & 5,826 & 1,908 & 5,198 & 7,106 & 1,594 & 1,031 & 2,625 \\
\# Answers & 19,360 & 500 & 305 & 1,009 & 1,107 & 380 & 292 & 438 \\
\Xhline{2\arrayrulewidth}
\end{tabular}
\begin{tablenotes}
    \small
    \item (*) PQL-3H-More has twice more QA pairs (2,062) with the same number of entities, relations, knowledge facts and answers as PQL-3H. 
\end{tablenotes}
\end{threeparttable}
\caption{Statistics of four benchmark datasets: Knowledge-aware Visual Question Answering (KVQA), Fact-based Visual Question Answering (FVQA), PathQuestion (PQ) and PathQuestion-Large (PQL).}
\label{table:datastat}
\end{table*}

\begin{table*}[h]
\centering
\begin{tabular}{l|cccccccccc}
\Xhline{2\arrayrulewidth}
 & Bool & Comp. & \begin{tabular}[c]{@{}c@{}}Multi\\ entity\end{tabular} & \begin{tabular}[c]{@{}c@{}}Multi\\ hop\end{tabular} & \begin{tabular}[c]{@{}c@{}}Multi\\ relation\end{tabular} & 1-hop & \begin{tabular}[c]{@{}c@{}}1-hop \\ subtract\end{tabular} & Spatial & Subtract. \\
 \hline
MemNN & 75.1 & 50.5 & 43.5 & 53.2 & 45.2 & 61.0 &  - &  48.1 & 40.5 \\
GCN & 86.8 & 87.7 & 87.7 & 96.7 & 77.7 & 61.4 & 53.7 & 29.4 & 37.7  \\
GGNN & 86.6 & 88.8 & 88.6 & 95.1 & 90.0 & 70.4 & 55.2 & 32.6 & 26.1 \\
HAN & 98.1 & 93.8 & 93.6 & 98.2 & 92.8 & 73.5 & 51.5 & 29.6 & 29.0 \\
BAN & 98.5 & 94.8 & 94.5 & \textbf{99.3} & 98.6 & 81.2 & 56.7 & 39.1 & 39.2 \\
\hline
\textbf{Ours} & \textbf{99.1} & \textbf{96.9} & \textbf{96.8} & 99.2 & \textbf{99.3} & \textbf{89.9} & \textbf{73.3} & \textbf{90.1} & \textbf{42.4} \\
 \Xhline{2\arrayrulewidth}
\end{tabular}
\caption{Analysis of QA accuracy over different question categories of original (ORG) questions in oracle setting. All models use 3-hop graph reported in Table \ref{table:orc}. Comp. and Subtract. are abbreviations of Comparison and Subtraction. The best performance of each question type is highlighted in bold.}
\label{table:qcate}
\end{table*}

\begin{figure*}[h]
  \centering
  \includegraphics[width=0.95\textwidth]{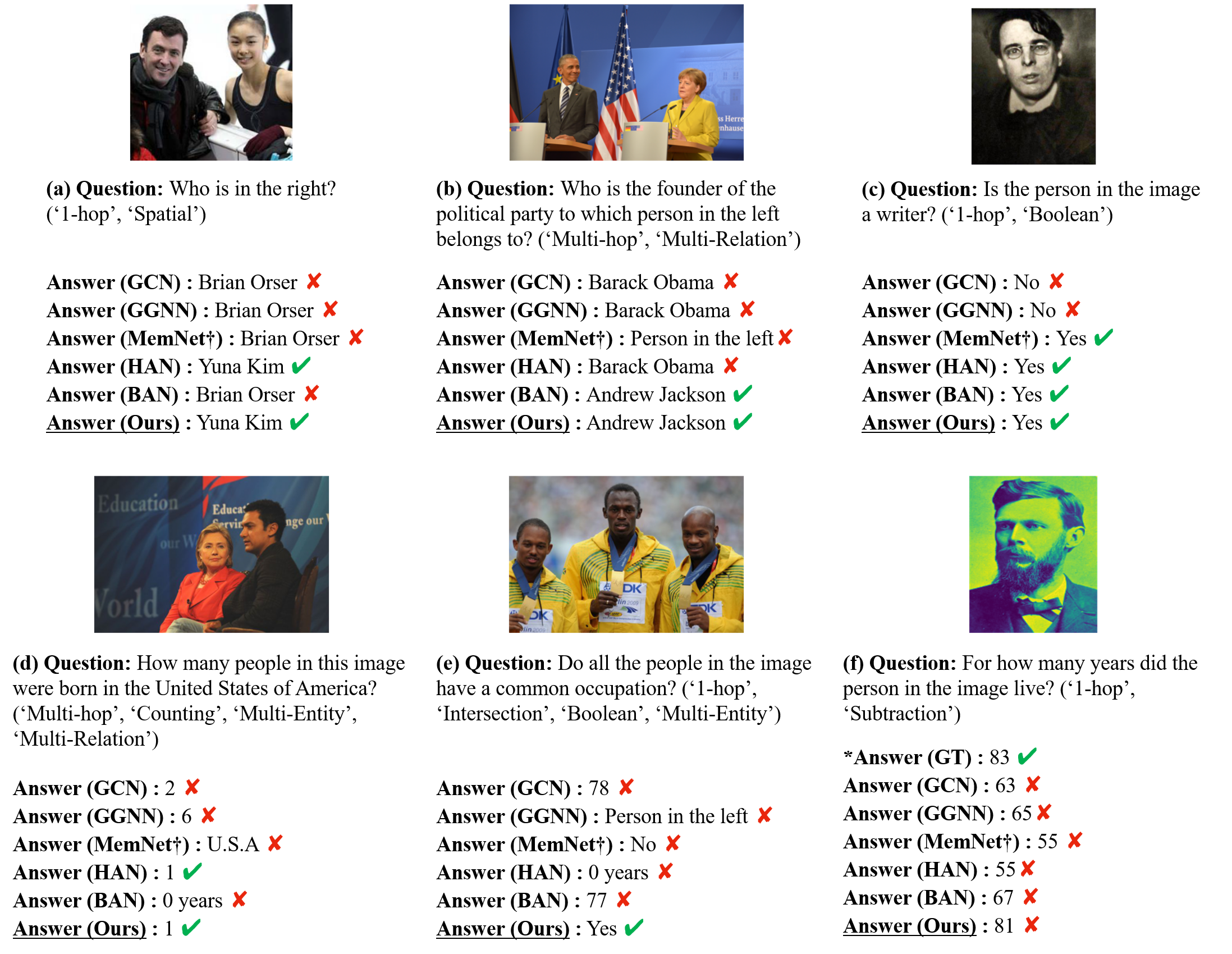}
  \caption{Qualitative results on KVQA dataset. GCN, GGNN, MemNN$\dagger$, HAN, BAN and our model infer answers to a question about a given image. Green and red marks indicate correct and incorrect answers, respectively.}
  \label{fig:qualitative}
\end{figure*}

\section{Additional Analysis on KVQA}\label{ax:kvqa}
Here, we analyze more in-depth on KVQA dataset concerning i) categories of question, and ii) types of answer selector. All models are under the same setting of ORG+3-hop reported in Table \ref{table:orc}.

\subsection{Analysis on question categories} We analyze QA performances over different question categories in Table \ref{table:qcate}. Hypergraph Transformer achieves the best accuracy in all categories except Multi-hop (slightly low at second-best). Our model shows notable strengths especially on complex problems such as Comparison, Multi-entity or Subtraction. To draw inferences for these question categories, the model needs to attend to multiple knowledge facts related to a given question, and conducts multi-hop reasoning based on the facts. Also, our model shows significant improvement in spatial question compared to other models. Whereas spatial question is quite simple, it is required to understand a correct spatial relationship between multiple entities in a given image. Examples of QA on diverse question categories are depicted in Figure \ref{fig:qualitative}. Answers, inferred by five comparative models and the proposed model, are presented with corresponding image and question. The qualitative results indicate that our model draws reasonable inferences across diverse question categories.

\begin{table*}[h]
\centering
\begin{tabular}{c|cccc|cccc}
\Xhline{2\arrayrulewidth}
\multicolumn{1}{l|}{} & \multicolumn{4}{c|}{Original (ORG)} & \multicolumn{4}{c}{Paraphrased (PRP)} \\
 & Zero-shot & One-shot & Multi-shot & ALL & Zero-shot & One-shot & Multi-shot & ALL \\
\hline
MLP & 0.0 & 78.3 & 87.2 & 76.0 & 0.0 & 76.9 & 86.8 & 75.6 \\
SIM & \textbf{93.9} & \textbf{96.7} & \textbf{90.1} & \textbf{91.0} & \textbf{92.4} & \textbf{96.3} & \textbf{89.9} & \textbf{90.7} \\
\Xhline{2\arrayrulewidth}
\end{tabular}
\caption{Analysis for answer selector with the frequency of answers in the test split. SIM and MLP represent similarity-based answer selector and multi-layer perceptron.}
\label{table:SIMvsFC}
\end{table*}

\subsection{Effect of similarity-based answer selector} To validate the impact of similarity-based answer selector, we replace the similarity-based answer selector (SIM) with a multi-layer perceptron (MLP). We first note that KVQA dataset includes a large number of unique answers (19,360), and contains a lot of zero-shot and few-shot answers in test phase. As shown in Table \ref{table:SIMvsFC}, the MLP fails to infer zero-shot answers which are not appeared in the training phase at all. Besides, the performance difference between SIM and MLP in one-shot answer (appeared in the only one time in training phase) is more than 18\%. The MLP uses 17\% more parameters than SIM because KVQA has a large number of answer candidates (19,360). When the number of candidate answers increases, the MLP needs more parameters, but SIM does not. To sum up, the similarity-based answer selector (SIM) contributes to infer few-shot and zero-shot answers in parameter-efficient manner.

\begin{figure*}[h]
  \centering
  \includegraphics[width=0.9\textwidth]{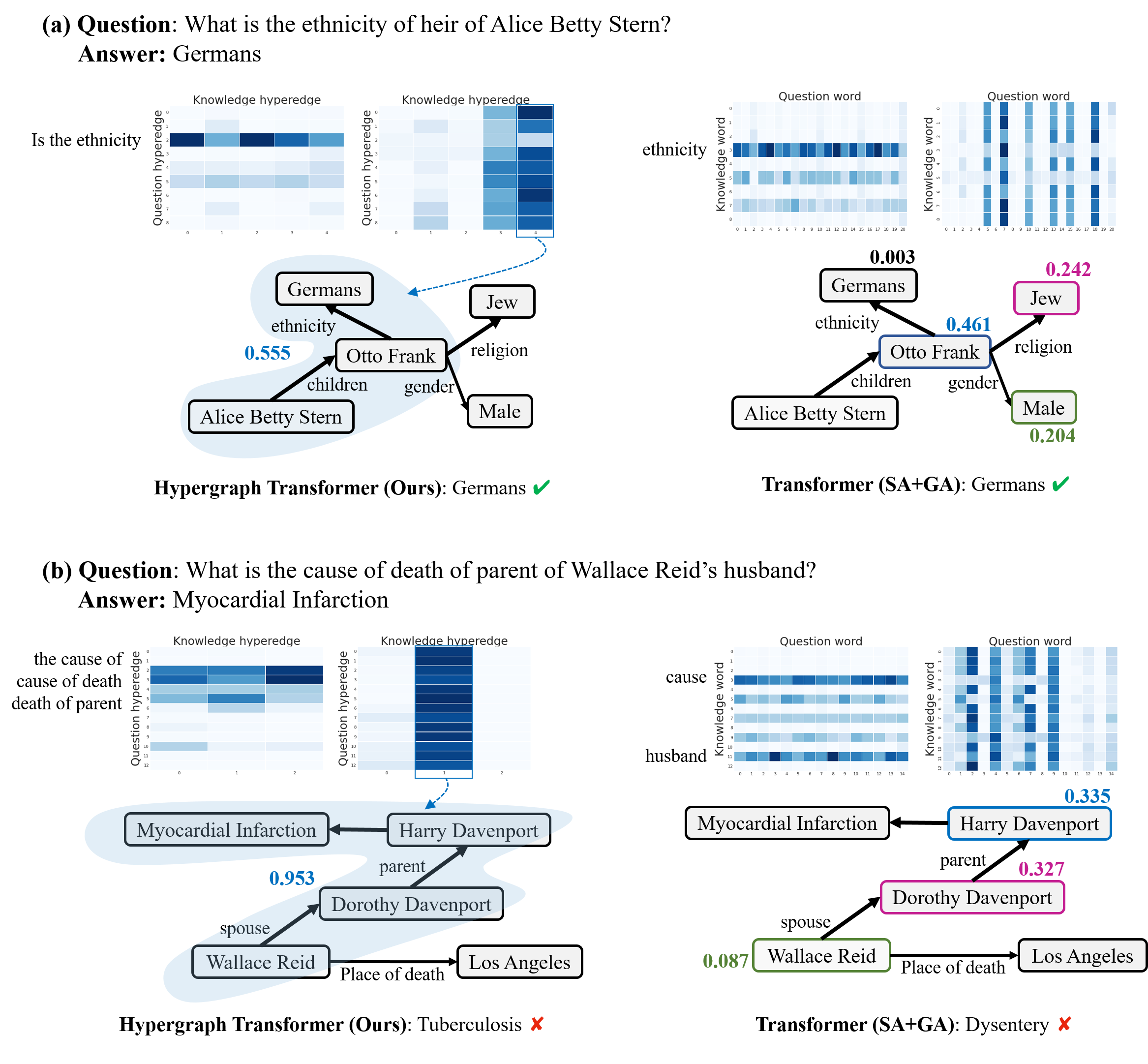}
  \caption{Qualitative analysis on effectiveness of using hypergraph as input format to Transformer architecture. Here, we visualize attention maps ($\operatorname{Attention}(Q_k, K_q, V_q)$ and $\operatorname{Attention}(Q_q, K_k, V_k)$) for Hypergraph Transformer and the Transformer (SA+GA). All attention scores are averaged over multi-heads and multi-layers. Each $x$ and $y$ axis represent indices of question and knowledge hyperedges in Hypergraph Transformer, and indices of question and knowledge word in Transformer (SA+GA). In the attention maps, the dark colors represent high values. We also visualize the top-3 attended knowledge hyperedges in Hypergraph Transformer, and top-3 attended knowledge fact in Transformer (SA+GA) with the attention score.}
  \label{fig:qual_PQnPQL}
\end{figure*}

\section{Qualitative Analysis on PathQuestion}\label{ax:pq}
Figure \ref{fig:qual_PQnPQL} shows the qualitative analysis of Hypergraph Transformer and Transformer (SA+GA) on PathQuestion. 
In Figure \ref{fig:qual_PQnPQL}(a), Hypergraph Transformer attends to the second question hyperedge $\{\textit{the} \preceq \textit{ethnicity} \preceq \textit{of}\}$ and the fourth knowledge hyperedge $\{\textit{Alice Betty Stern} \preceq \textit{children} \preceq \textit{Otto Frank} \preceq \textit{ethnicity}\preceq \textit{Germans}\}$ to reason based on the multi-hop evidence about ethnicity.
On the other hand, Transformer (SA+GA) focuses on the third question word \textit{ethnicity} correctly, but attends to \textit{Otto Frank}, \textit{Jew}, \textit{Male} with the high attention score 0.461, 0.242, and 0.204, not the exact knowledge entity, \textit{Germans}. In Figure \ref{fig:qual_PQnPQL}(b), both model, Hypergraph Transformer and Transformer (SA+GA), fail to infer the correct answer. The predicted answer of Hypergraph Transformer is wrong even though it attends correctly to the first knowledge hyperedge $\{\textit{Wallace Reid} \preceq \textit{spouse} \preceq \textit{Dorothy Davenport} \preceq \textit{parents} \preceq \textit{Harry Davenport} \preceq \textit{cause of death} \preceq \textit{Myocardial Infarction}\}$. However, Transformer (SA+GA) attends to only the second and seventh word (\textit{Dorothy Davenport}) and the fourth and ninth word (\textit{Harry Davenport}) in knowledge with high attention score, not the answer entity, \textit{Myocardial Infarction}. We consider that the reason why Hypergraph Transformer failed to infer the correct answer despite focusing on the exact knowledge fact is that the correct answer word (\textit{Myocardial Infarction}) appears rarely in QA pairs. 

\section{Experimental details}\label{ax:expdetails}

\subsection{Knowledge-aware VQA}
We follow the experimental settings suggested in~\cite{shah2019kvqa}. For entity linking, we apply well-known pre-trained models for face identification: RetinaFace~\cite{deng2020retinaface} for face detection and ArcFace~\cite{Deng_2019_CVPR} for face feature extraction. We first assign a name of the detected faces with the label of the closest distance compared to all of the face embeddings of 18,880 named entities. In addition, we refine a list of detected named entities by matching the associated image caption (i.e., Wikipedia caption). By doing so, we obtain the result of entity linking with top-1 precision 65.0\% and top-1 recall 72.8\%. QA performances in the entity linking setting on KVQA are shown in Table \ref{table:det}. Here, we note that BLSTM and MemNN of the first section in the table are based on the different entity linking modules with top-1 precision 81.1\% and top-1 recall 82.2\%\footnote{The code for the entity linking module has not been released publicly. As such, we implement the module based on the open-source: https://github.com/deepinsight/insightface. We use the pre-trained model named retinaface-mnet025-v2 and LResNet100E-IR,ArcFace@ms1m-refine-v2.}. It is more accurate than ours around 9.4\% in the recall metric. 

\begin{table}[]
\centering
\begin{tabular}{l|cc|c}
\Xhline{2\arrayrulewidth}
    Model & ORG & PRP & Mean \\
    \hline
    BLSTM & 48.0 & 27.2 & 37.6 \\
    MemNN & 50.2 & 34.2 & 42.2 \\
    \hline
    GCN & 48.9 & 48.2 & 48.5 \\
    GGNN & 50.9 & 50.9 & 50.9 \\
    MemNN$\dagger$ & 54.0 & 53.9 & 54.0 \\
    HAN & 53.4 & 53.3 & 53.3 \\
    BAN & 59.6 & 60.0 & 59.8 \\
    Transformer (SA) & 57.5 & 58.9 & 58.3 \\
    Transformer (SA+GA) & 60.4 & 59.8 & 60.1 \\
    \hline
    \textbf{Ours} & \textbf{62.0} & \textbf{62.8} & \textbf{62.4}\\
\Xhline{2\arrayrulewidth}
\end{tabular}
\caption{QA accuracy on entity linking setting in KVQA. The performances of BLSTM and MemNN are reported in~\cite{shah2019kvqa}.} 
\label{table:det}
\end{table}

\begin{table}[]
\centering
\begin{tabular}{lcc}
\Xhline{2\arrayrulewidth}
     & \multicolumn{2}{c}{Accuracy} \\
     & @1 & @3 \\
\hline
    Human & 77.99 & - \\
\hline
    LSTM-Q+I (Pre-VQA) & 24.98 & 30.30 \\
    Hie-Q+I (Pre-VQA) & 43.14 & 59.44 \\
    FVQA-Top3-QQmaping & 56.91 & 64.65 \\
    STTF-Q+VConcept & 62.20 & 75.60 \\
    RC (pre-SQuAD) & 62.94 & 70.08 \\
    Out of the Box	& 69.35	& 80.25 \\
    Mucko & 73.06 & \textbf{85.94} \\
    \hline
    \textbf{Ours} & \textbf{76.55} & 82.20 \\ 
\Xhline{2\arrayrulewidth}
\end{tabular}
\caption{Accuracy on Fact-based Visual Question Answering (FVQA). Top-1 and top-3 accuracy are used as evaluation metrics.}
\label{table:fvqa}
\end{table}

\subsection{Fact-based VQA}
We follow the experimental settings suggested in~\cite{wang2018fvqa}. Following the paper, the dataset provides five splits of train and test data. We report the average accuracy of five repeated runs on different data split: 76.55 as top-1 accuracy (average of 76.93, 75.92, 76.24, 76.16, and 77.50) and 82.20 as top-3 accuracy (average of 82.90, 81.45, 81.70, 81.74 and 83.20). The experimental results are shown in Table \ref{table:fvqa}.

\subsection{PathQuestion and PathQuestion-Large}
We follow the same experimental settings suggested in~\cite{zhou2018pqpql}. Following the paper, we split the dataset into train, validation, and test sets with a proportion of 8:1:1, and report the average accuracy of five repeated runs on different data split. 

\section{Implementation Details of Comparative Models for KVQA}\label{ax:impdetailsKVQA}
For comparative models for KVQA, three kinds of methods are considered, which are graph-based, memory-based and attention-based networks.

\paragraph{Graph-based networks.} Graph convolutional networks (GCN)~\cite{kipf2017semi} and gated graph neural networks (GGNN)~\cite{ggnn} are representative models of graph-based neural networks. Both learn node representations of a knowledge and question graph (not a hypergraph), propagating information between neighborhoods. After propagation, node representations in a graph are aggregated to encode a graph-level representation. Joint representation is obtained based on the two graph representations. 
\paragraph{Memory-based networks.} Memory network (MemNN)~\cite{memnet} is a de facto baseline for fact-based question answering. Each fact is embedded into a memory slot, and soft attention is calculated between memory slots and a given question. Joint representation is obtained based on the attention.
\paragraph{Attention-based networks.} Bilinear attention networks (BAN)~\cite{kim2018bilinear} and hypergraph attention networks (HAN)~\cite{kim2020hypergraph} consider interactions between knowledge and question based on co-attention mechanism. BAN calculates soft attention scores between knowledge entities and question words. Meanwhile, HAN employs stochastic graph walk in a knowledge and question graph to encode high-order semantics (e.g., knowledge facts and question phrases), and considers attention scores between knowledge facts and question phrases. Joint representation is obtained based on the attention as well. The more implementation details of the above comparative models is described as follows.

\subsection{Graph convolutional networks}
The knowledge and question graph are encoded separately by two graph convolutional networks (GCN)~\cite{kipf2017semi}. Each GCN model consists of two propagation layers and a sum pooling layer across the nodes in the graph. The operation of the propagation layer is as follows: $f(H^{(l)}, A)= \sigma(\hat{D}^{-\frac{1}{2}} \hat{A} \hat{D}^{-\frac{1}{2}} H^{(l)} W^{(l)})$ where $\hat{A}=A+I$, $A$ is an adjacency matrix of the graph, $I$ is an identity matrix, $D$ is a degree matrix of $A$, $W^{(l)}$ is the model parameters of $l$-th layer, and $H^{(l)}$ is the representations of the graph in the $l$-th layer. Here, $H^{(0)}$ is the word embeddings of each entity in the knowledge and question graph. After propagation and aggregation phase, the knowledge and question graph representations are obtained. Then, the two graph representations are concatenated and fed into a single layer feed-forward layer to get joint representation. 

\subsection{Gated graph neural networks}
As the same as graph convolutional networks, the knowledge and question graph are encoded separately by two gated graph neural networks (GGNN). Each GGNN model consists of three gated recurrent propagation layers and a graph-level aggregator. Motivated by Gated Recurrent Units~\cite{cho2014gpu}, GGNN adopts a update gate and a reset gate to renew each node's hidden state. The detailed equation of gated recurrent propagation is as follows: $\mathbf{h}_v^{(1)} = [\mathbf{x}_v^{T},\mathbf{0}]^T$ where $\mathbf{x}_v$ is the $v$-th word embedding of each entity in the knowledge and question graph, $\mathbf{a}_v^{(t)} = A_{v:}^T \ [\mathbf{h}_1^{(t-1)^T} \cdots \mathbf{h}_{|\mathcal{V}|}^{(t-1)^T}]^T + \mathbf{b}$ where the matrix $A$ determines how nodes in the graph communicate each other and $\mathbf{b}$ is a bias vector. Then, the update gate and reset gate are computed as follows: $\mathbf{z}_v^{t} = \sigma(W^{z}\mathbf{a}_v^{(t)} + U^{z}\mathbf{h}_v^{(t-1)}), \ \ \mathbf{r}_v^t = \sigma(W^{r}\mathbf{a}_v^{(t)} + U^{r}\mathbf{h}_v^{(t-1)})$ where $\sigma$ is a logistic sigmoid function, and $W^{[\cdot]}$ and $U^{[\cdot]}$ are learnable parameters. Finally, the hidden states of nodes in the given graph are updates as $ \mathbf{h}_v^{(t)} = (1-\mathbf{z}_v^t) \odot \mathbf{h}_v^{(t-1)} + \mathbf{z}_v^t \odot \tilde{\mathbf{h}}_v^{(t)}$ where $\tilde{\mathbf{h}}_v^{(t)} = \tanh({W^h\mathbf{a}_v^{(t)} + U^h(\mathbf{r}_v^t \odot \mathbf{h}_v^{(t-1)}})$. After the propagation phase, the nodes in the graph are aggregated to a graph-level representation as $\mathbf{h}_\mathcal{G} = \tanh(\sum_{v\in\mathcal{V}}\sigma(i(\mathbf{h}_v^{(T)},\mathbf{x}_v)) \odot \tanh(j(\mathbf{h}_v^{(T)},\mathbf{x}_v))$ where $i$ and $j$ are a single layer feed-forward layer, respectively. Then, the two aggregated graph representations are concatenated and fed into another single layer feed-forward layer to get joint representation of question and knowledge graph. 

\subsection{Memory networks}
We reproduce end-to-end memory networks~\cite{sukhbaatar2015end} proposed as a baseline model in~\cite{shah2019kvqa}. First, we use Bag-of-words (BoW) representation for knowledge facts and a question. The soft attention over the knowledge facts and the given question is computed as follows: $p_{ij}=\operatorname{softmax}(q_{i-1}^Tm_{ij})$ where $m$ is the embeddings of knowledge facts, $i$ is a number of layer and $j$ is an index of knowledge facts. The output representation of $i$-th layer is $O_{i}=\sum_j p_{ij}o_{ij}$ where $o$ is the another embeddings of knowledge facts different from $m$. The updated question representation is $q_{k+1}=O_{k+1}+q_k$, and based on the output representation and question representation, answer is predicted as follows: $\hat{a}=\operatorname{softmax}(f(O_K+q_{K-1}))$ where $f$ is a single layer feed-forward layer. Here, we set up the model as three layers with adjacent and layer-wise weight tying. 

\subsection{Bilinear attention networks}
Bilinear attention networks exploit a multi-head co-attention mechanism between knowledge and question. BAN calculates soft attention scores between knowledge entities and question words as follows: $\mathcal{A}=\operatorname{softmax}(W^h \circ (M^qW^q){(M^kW^k)}^\top)$ where $M^q, M^k$ are a row-wise concatenated question words and knowledge entities, $W^{[\cdot]}$ is learnable matrices, and $\circ$ is element-wise multiplication. Based on the attention map $\mathcal{A}$, the joint feature is obtained as follows: $z_i={(M^qW^q)_i}^\top \mathcal{A} (M^kW^k)_i$ where the subscript $i$ denotes the $i$-th index of column vectors in each matrix. For multi-head attention, the attended outputs with different heads are concatenated and fed into a single layer feed-forward layer to make a final representation. Here, we use four attention heads as multi-head.

\subsection{Hypergraph attention networks}
The model architecture and detailed operation of hypergraph attention networks are similar to that of BAN. The difference between BAN and HAN is the abstraction level of the input. For HAN, the hyperedges sampled by stochastic graph walk are fed into the co-attention mechanism. What HAN and our model have in common is introducing a hypergraph to consider high-order relationships in question graph and knowledge graph. Both models share the similar motivation, but the core operations are quite different. Especially, HAN employs stochastic graph walk to construct question and knowledge hypergraph. Due to the randomness of the stochasticity, misinformed or incomplete hyperedges can be extracted.

\subsection{Transformer Variants}
The model architectures of Transformer (SA) and Transformer (SA+GA) presented in this paper are the same as Hypergraph Transformer. The only difference is the abstraction level of input. The Transformer (SA) and Transformer (SA+GA) take single-word-unit as input tokens, and Hypergraph Transformer takes hyperedges as input tokens. Following~\cite{vaswani2017attention,tsai2019multimodal}, we apply positional embeddings to the input sequence of both models. We stack two guided-attention blocks and three self-attention blocks, respectively. Each attention block has multi-head attention with four attention heads followed by layer normalization, residual connections and a single multi-layer perceptron. We set the dropout applied on the token embedding weights, query and key-value embedding weights, attention weights and residual connections from 0.05 to 0.2. We minimize negative log-likelihood using Adam optimizer~\cite{iclr15adam} with an initial learning rate from $1e-4$ to $1e-5$ with batch size from $128$ to $256$. All transformer variant models described in this paper have the same fixed-number of sequence length as follows: 300 for 1-hop, 1,000 for 2-hop and 1,800 for 3-hop graphs.

\end{document}